\documentclass{bmvc2k}
\usepackage{amssymb}
\usepackage{pifont}
\usepackage{multirow}
\usepackage[font=small,labelfont=bf,labelsep=colon]{caption}
\usepackage{amsmath}
\usepackage{wrapfig}
\usepackage[linesnumbered,ruled]{algorithm2e}
\SetKwProg{Init}{init}{}{}
\SetArgSty{textup}
\usepackage{pgfplots}

\newcommand{\myparagraph}[1]{\vspace{2pt}\noindent{\bf #1}}

\title{Human Attention in Fine-grained Classification}

\addauthor{Yao Rong}{yao.rong@uni-tuebingen.de}{1}
\addauthor{Wenjia Xu}{xuwenjia16@mails.ucas.ac.cn}{2}
\addauthor{Zeynep Akata}{zeynep.akata@uni-tuebingen.de}{1,3}
\addauthor{Enkelejda Kasneci}{enkelejda.kasneci@uni-tuebingen.de}{1}

\addinstitution{
University of T{\"u}bingen, Germany
}
\addinstitution{
University of Chinese Academy of Sciences, Beijing, China
}
\addinstitution{
Max Planck Institute for Intelligent Systems, T{\"u}bingen, Germany
}

\runninghead{Rong et al.}{Human Attention in Fine-grained Classification}

\begin{document}
\maketitle

\begin{abstract}
The way humans attend to, process and classify a given image has the potential to vastly benefit the performance of deep learning models. Exploiting where humans are focusing can rectify models when they are deviating from essential features for correct decisions. To validate that human attention contains valuable information for decision-making processes such as fine-grained classification, we compare human attention and model explanations in discovering important features. Towards this goal, we collect human gaze data for the fine-grained classification dataset CUB and build a dataset named CUB-GHA (Gaze-based Human Attention). Furthermore, we propose the Gaze Augmentation Training (GAT) and Knowledge Fusion Network (KFN) to integrate human gaze knowledge into classification models. We implement our proposals in CUB-GHA and the recently released medical dataset CXR-Eye of chest X-ray images, which includes gaze data collected from a radiologist. Our result reveals that integrating human attention knowledge benefits classification effectively, e.g. improving the baseline by 4.38\% on CXR. Hence, our work provides not only valuable insights into understanding human attention in fine-grained classification, but also contributes to future research in integrating human gaze with computer vision tasks. CUB-GHA and code are available at \url{https://github.com/yaorong0921/CUB-GHA}.
\end{abstract}

\section{Introduction}
\begin{figure*}[t]
\begin{center}
   \includegraphics[width=\linewidth]{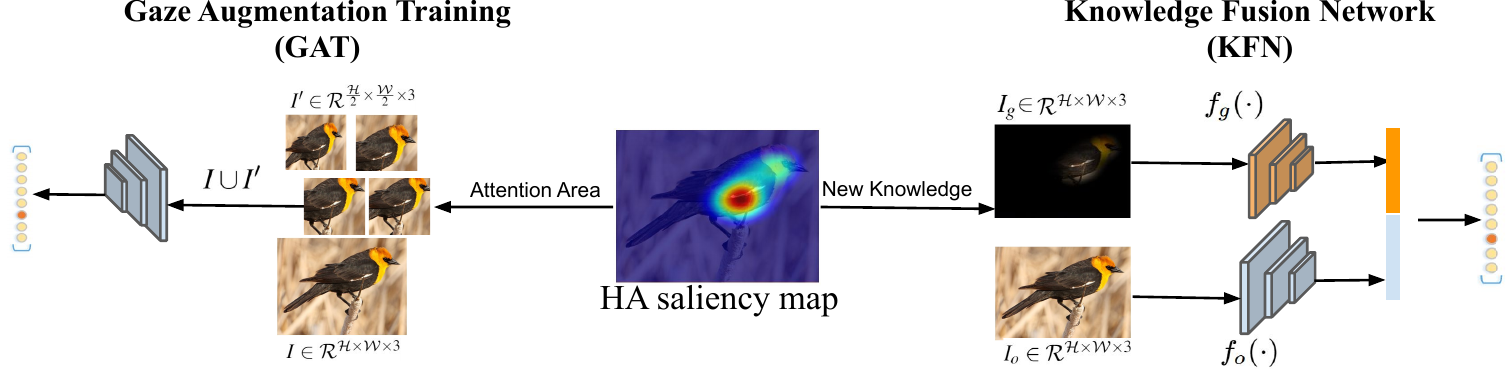}
\end{center}
   \caption{Overview of our proposed methodology. HA salicency map is used to obtain attention area which is used to enhance the training dataset in Gaze Augmentation Training (Left), while it is used as extra knowledge and fused together with the image knowledge in the Knowledge Fusion Network (Right).
   }
   \vspace{-5mm}
\label{fig:model}
\end{figure*}

Through a lifelong learning process, humans have developed a selective attentional mechanism, which has received attention in many areas of artificial intelligence \cite{zhang2020human}. As human attention can be revealed from gaze data, it bears the potential to explain our behavior and decisions~\cite{posner1990attention}. 
Many computer vision applications embrace human gaze information to detect salient objects for solving tasks \cite{shanmuga2015eye,santella2006gaze,Karessli_2017_CVPR}. To visually illustrate human attention in these tasks, it is common to add a Gaussian filter on fixation points to form a feature map \cite{Judd_2012}, which is also called \textit{saliency} map \cite{kummerer2016deepgaze}~(see Figure~\ref{fig:model}). 
Similar to how gaze explains human decisions, the post-hoc attention of a network, i.e. model explanation, tries to reveal important regions for neural network decision-making \cite{18_rise,19_cam,20_grad_cam,IG,shrikumar2016not,hooker2018benchmark}. Both can be visualized by means of saliency maps, thus allowing the study of similarities and differences between them. In this context, several previous works show that humans and models are looking at different regions when performing the same task \cite{das2017human,sen2020human}. However, it is not clear whether a feature discovered by a human is more efficient for solving a given task or not. Our work addresses this research gap and the hypotheses that (1) human attention focuses on essential features for solving the task (e.g. fine-grained classification); (2) using human attention also allows improving model performance in accomplishing the task. To validate the first hypothesis, we first capture and present human attention in the style of a saliency map. We compare the regions that human attention covers with the ones that are discovered by the model (model explanation), and show that human attention hints on the regions that are more discriminative in the classification. We propose two modules which make use of the essential features revealed by human gaze to validate the second hypothesis: we use Gaze Augmentation Training (GAT) to train a better classifier and a Knowledge Fusion Network (KFN) to integrate the human attention knowledge into models.

Our contributions are as follows: (1) We collect human gaze data for the fine-grained data set CUB, enhance it by incorporating human attention and coin this new dataset as \textbf{CUB-GHA} (Gazed-based Human Attention). For this novel dataset, we also validate the efficiency of human gaze data in discovering discriminative features. (2) We propose two novel modules to incorporate human attention knowledge in classification tasks: Gaze Augmentation Training (GAT) and Knowledge Fusion Network (KFN). (3) To showcase the relevance of our work for highly relevant applications, we evaluate our methods not only on our novel CUB-GHA dataset, but also on chest radiograph images from a recently released dataset CXR-Eye (which contains also gaze data). Our work shows that human attention knowledge can be successfully integrated in classification models and help improve the model performance with regard to the state-of-the-art in different classification tasks.

\section{Related Work}
\label{sec:relatedWork}
\myparagraph{Human Gaze in Machine Learning.}
Recent developments in hardware devices allow for the precise recording of eye movements in different activities, ranging from human-computer interaction \cite{mathe2014actions, majaranta2014eye} to complex and dynamic real-world tasks, such as driving \cite{xia2020periphery, braunagel2017ready} and robotics \cite{weberdistilling, shafti2019gaze,aronson2018eye}. Furthermore, the way that visual information is processed can reveal information about a person's strategy or level of expertise~\cite{castner2020deep}. In the medical domain, researchers have validated that gaze data reveals patterns which can benefit AI models, as for disease (Pneumonia and Congestive Heart Failure) classification \cite{karargyris2021creation}.  
In computer vision, gaze data has proven its usefulness in various applications \cite{santella2006gaze,shanmuga2015eye,Karessli_2017_CVPR,qiuxia2020understanding}. E.g., \cite{Karessli_2017_CVPR} collects gaze (coordinates, duration, etc.) vectors for 60 bird classes in dataset \cite{WelinderEtal2010} to form embeddings for zero-shot learning. \cite{qiuxia2020understanding} compares the attention map generated by an attention module (two convolutional layers) with human attention maps generated by the data from \cite{Karessli_2017_CVPR} and shows that human attention surpasses the attention module. \cite{santella2006gaze} proposes a photograph cropping system using the collected fixation data to identify important content and compute the best crop. Eye tracking data is also used to extract dominant objects in videos \cite{shanmuga2015eye}. Different from previous works which use gaze for specific tasks \cite{santella2006gaze,shanmuga2015eye,Karessli_2017_CVPR}, our proposal GAT leverages human attention to train a better backbone which can be used in many different tasks and frameworks. Moreover, we evaluate GAT and KFN for two different classification tasks and thus show the general validity of our methods.
 

\myparagraph{Attention Module in Fine-grained Classification.} Many previous works \cite{zheng2017learning, sun2018multi, liu2016fully,zhang2021multi,liu2017localizing,li2017dynamic,sermanet2014attention,fu2017look,zheng2019looking,zhuang2020learning,ji2020attention} integrate attention modules in networks to localize the parts which are important for fine-grained classifications and make use of the information of the discriminative parts to improve the models' performance.
\cite{liu2017localizing,sermanet2014attention,liu2016fully,li2017dynamic,fu2017look} adopt the Recurrent Attention Model (RAM) \cite{mnih2014recurrent}, where an attention agent is deployed to predict locations of the discriminative regions, and train the classifier based on these cropped regions. The attention agent is trained with a reinforcement learning algorithm to address the non-differentiability due to the cropping operation. However, the architecture of this attention model is cumbersome with high computational cost.
\cite{zheng2017learning, sun2018multi,zhang2021multi,zheng2019looking, zhuang2020learning, ji2020attention}, on the other hand, design attention modules using the output from intermediate layers in networks and enforce it to capture discriminative features. Compared to previous works, we do not use the intermediate outputs from networks to generate model attention but use human attention maps. Our method augments the training set with regions cropped according to human attention and thus accomplishes training a better classifier. We compare our method with previous works and demonstrate the profit of exploiting human attention in Section \ref{sec:experiments}.

\section{CUB-GHA Dataset}
\label{sec:dataset}
In this section, we first provide the details of our gaze data collection paradigm and then analyze the effect of machine explanation and human attention to the fine-grained classification model. 
To collect gaze data, we employ the CUB-200-2011 (CUB)~\cite{WahCUB_200_2011} dataset with 11,788 images from 200 bird classes incorporating various annotations: image-level attributes, body part locations, and text descriptions of the bird. Our annotation leads to a human-gaze enhanced version, i.e. CUB-GHA.

We choose the fine-grained CUB dataset for two reasons: 1) The difference between two similar classes lies in local and compositional attributes, which can be precisely captured by human gaze. For instance, it is challenging to achieve a measure for unified human attention when comparing a bear and a horse as there are many differences between them. In contrast, distinguishing between two similar birds with different throat colors presents a more unified problem~(as shown in Figure~\ref{fig:setup}). 2) The CUB dataset is widely used for various computer vision tasks, such as fine-grained classification~\cite{fu2017look,zheng2019looking,dubey2018maximum}, zero-shot learning~\cite{ALE,DEM,xian2018zero,xu2020attribute}, explainable artificial intelligence~\cite{kanehira2019learning,chen2019looks,anne2018grounding}, etc. Thus, our CUB-GHA may serve as a valuable foundation for exploring the effect of human attention on those tasks. 

\subsection{Gaze Data Collection}
\label{sec:HA maps}
\begin{figure*}[]
\begin{center}
   \includegraphics[width=\linewidth]{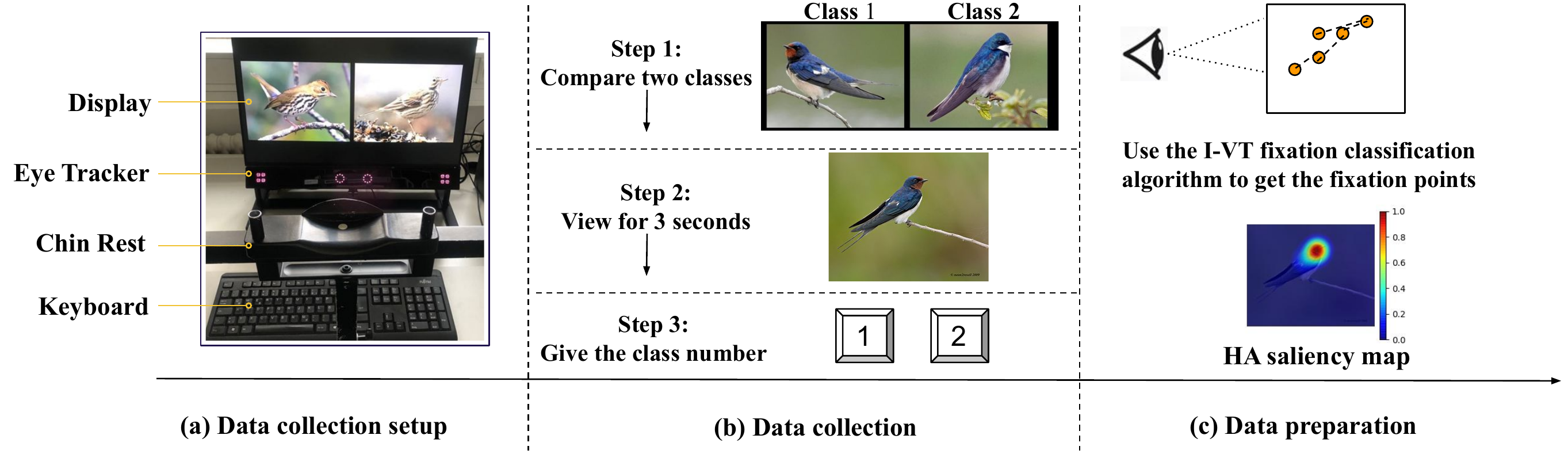}
\end{center}
   \caption{\textbf{(a)} Eye tracker set-up: We use a Tobii Spectrum eye-tracker to capture gaze information at a high frequency of 1200 Hz. \textbf{(b)} Data collection: Step 1 represents a schematic overview of the image comparison task where two images of different species are freely viewed. In Step 2, a randomly selected example of one of the species is shown to the user for which gaze data is then collected. To gamify this setting, the user is asked to choose the correct class in Step 3. \textbf{(c)} Preparing human attention data: we visualize human attention in Gaussian-based saliency maps. 
   }
   \vspace{-5mm}
\label{fig:setup}
\end{figure*}

\myparagraph{Collection Framework.} As illustrated in \cite{Karessli_2017_CVPR}, humans fixate on class-discriminative features when they observe two very similar classes. In this paper, we adopt an image comparison game~\cite{Karessli_2017_CVPR}, where we encourage participants to look at the discriminative features when comparing two similar images from different categories. The comparison task is designed to be challenging to provide more powerful insights, i.e. two classes in one comparison pair are chosen to be very similar. 

A schematic overview of our data collection is presented in Figure \ref{fig:setup}. Figure \ref{fig:setup} (a) shows the experimental setup including a picture of the eye-tracker (Tobii Spectrum Eye Tracker, sampling at 1200 Hz) and the chin rest as well as the display (1920 $\times$ 1080 resolution). The chin rest is used to ensure precise recordings of the eye movements. Each image is re-scaled to fit to the screen and placed at the center. The average distance between the participant's nose and the screen is approximately 60 $cm$. The comparison task consists of three steps shown in Figure \ref{fig:setup} (b). In step 1, we present two representative images at the same time, each from one bird class of the CUB dataset, e.g. representative images of Barn Swallow and Tree Swallow. We choose the comparison pairs under the same sub-classes, and then different persons manually check the visual similarity to make sure that the comparison is not too simple. The participants are allowed to observe the images for as long as they want. 
When the participant is ready for the classification task, in step 2, an image from one of the two classes of the CUB dataset is shown. The participant has to choose which category the image belongs to by viewing the image. Note that the image shown for classification is displayed for only 3 seconds to avoid explorative gaze behavior unrelated to the task. One collection session includes one image from each class, meaning that there are 200 images reviewed per session. 
Every image in CUB is reviewed by five different participants. 25 subjects (19 males and 6 females with mean age 27.64 $\pm$ 4.15) participate in the experiment. Although the participants do not take part in the same number of sessions and instances, we make sure that every participant views all classes in every session. It is worth noting that all participants are domain novices with no specific knowledge about birds.

\myparagraph{Gaze Data Preparation.} The raw gaze data is preprocessed to extract fixation locations using the Velocity-Threshold Identification (I-VT) algorithm \cite{olsen2012tobii}. The resulting fixation points offered in the dataset include coordinates and duration information. Based on this information, we generate saliency maps for human gaze as shown in Figure \ref{fig:setup} (c). Every fixation location is modelled as a Gaussian distribution ${G}(\mu,\,\sigma^{2})$, where $\sigma$ is 75 pixels (in the displace resolution), according to the ratio of the distance to the screen and the approximate foveal area of 2$^{\circ}$. The duration of the fixation is then used as a weight for its Gaussian distribution. Finally, the saliency map is presented in grayscale image form. From here on, we note the human attention saliency map generated from gaze data as HA.

\subsection{Gaze Data Analysis} 
In this section, we validate the hypothesis that \textit{HA covers discriminative regions for the fine-grained classification}. Given the same image and the same (visual) task, HA and model explanation (ME) reveal regions which are important in making decisions for humans and models, respectively. Thus, we compare HA with four MEs provided by a trained classifier (vanilla ResNet-50 \cite{he2016deep}) with a classification score of 85.58\% on CUB
, and validate that HA is able to discover features that better differentiate the bird from other bird classes. The four ME used are Class Activations Maps (CAM)~\cite{19_cam}, Gradient-based CAM (Grad-CAM)~\cite{20_grad_cam}, InputXGradient (IxG)~\cite{shrikumar2016not}, and IntegratedGradients (IG)~\cite{IG}.

\begin{wrapfigure}{r}{0.5\textwidth}
  \begin{center}
    \vspace*{-0.5cm}
    \includegraphics[width=0.5\textwidth]{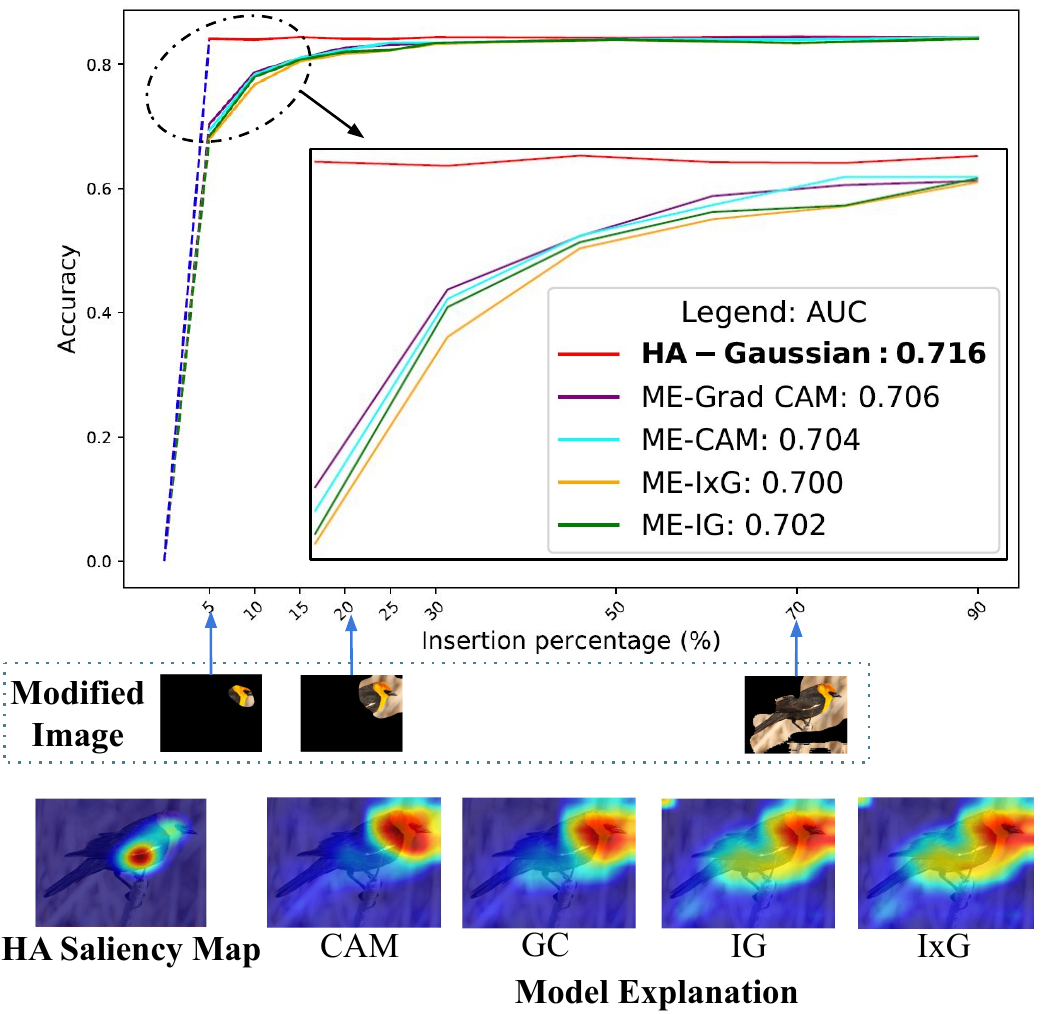}
  \end{center}
  \caption{Comparison of HA and ME in discriminative feature discovery. \textbf{Top:} Test accuracy on modified datasets using different saliency maps. The x-axis is the insertion percentage and the y-axis is the accuracy on test set. The AUC of each curve is reported in zoom-in image. \textbf{Middle:} modified images (using Grad-CAM as an example). \textbf{Bottom:} Illustration of HA and four MEs.}
  \label{fig:kar}
  \vspace*{-.4cm}
\end{wrapfigure}

For quantitative comparison, we compare HA and ME using the keep and retrain (KAR) procedure (proposed in the appendix to \cite{hooker2018benchmark}) to validate if the important regions highlighted by HA and ME help the model to make decisions. Concretely, we gradually insert important pixels to a blank image according to their values in HA or ME saliency maps. The modified percentage of pixels is [5,10,15,20,25,30,50,70,90]. After a certain amount of pixels are inserted, we retrain a new model using the modified train images and report the accuracy on modified test sets. Modified images at 5\%, 20\% and 70\% of pixels inserted using Grad-CAM are shown in Figure.\ \ref{fig:kar} (middle).
The intuition behind this is that the class-discriminative information should be included in the pixels that are evaluated as very important; with more pixels inserted which are relatively less important, the model performance will not improve much. If a saliency map selects the informative features as being the important ones for classification, the increase of accuracy at the beginning of insertion is rapid, i.e. the resulting higher Area Under the Curve (AUC) indicates a better feature importance estimate. 

The keep and retrain curves and the AUC scores for each method are shown in Figure.\ \ref{fig:kar} (top), and the qualitative saliency maps for HA and four MEs for one image are shown in the bottom. We see that HA and MEs do not focus on the same image regions: humans consider the white feathers on the black wing as a more important feature, while the model uses the yellow head as the most important feature (see the original image in Figure \ref{fig:model}). HA discovers more informative and important features for the fine-grained classification model than the MEs do, e.g. HA obtains an AUC score of 0.716 compared to Grad-CAM~(0.706) and IG~(0.702). With only $5\%$ important pixels revealed, the model trained with HA modified images can reach an accuracy of $81\%$ while the model trained with ME modified images only reaches an accuracy of around $70\%$. 
More details of the analyses can be found in the supplementary material.

\section{Methodology}
In this section, we introduce how we incorporate the gaze information to improve the classification performance, i.e. using gaze to augment training data (GAT) or as an extra information source (KFN). The illustration of the architecture is shown in Figure \ref{fig:model}.

\subsection{Gaze Augmentation Training}
Motivated by the assumption that the model should pay attention to the discriminative image regions~(highlighted by HA), we enhance our model's reaction to those regions by adding them as augmentation in training as illustrated in Figure \ref{fig:model} (left).


To get the $k$ augmentation images for the input image $I\in \mathcal{R}^{\mathcal{H} \times \mathcal{W} \times 3}$ (where $\mathcal{H}$ and $\mathcal{W}$ represent the width and height of the input image), we implement a sliding window algorithm to find areas which contain human attention. A window with the size of $(w,h)$ slides on the HA map $A \in \mathcal{R}^{\mathcal{H} \times \mathcal{W} \times 1}$ from the upper left to the right bottom corner (with stride size $s$ in both dimensions). We rank all the window areas according to the averaged pixel values inside windows and get $k$ cropped images according to top-$k$ highest scores. We resize the cropped images to the half of the width and height of the $I$, i.e. $I'\in \mathcal{R}^{\frac{\mathcal{H}}{2} \times \frac{\mathcal{W}}{2} \times 3}$, as suggested in \cite{sermanet2014attention,fu2017look, zheng2017learning} where the attended regions are resized into smaller sizes.
$I'$ has the same label $y$ as $I$ does. To get various regions, we use various window sizes and the non-maximum suppression. The training set is extended to $I \cup I'$. We train the model on the enlarged dataset with cross-entropy loss. Note that GAT just needs human gaze information in training and the model takes only original images as inputs in the test phase. 

\subsection{Knowledge Fusion Network}
\label{sec:KFN}
As shown in Fig.\ \ref{fig:model} (right), our KFN is a two-branch network that fuses the knowledge from HA and the original image features together.
The first branch is the image knowledge branch. This branch takes the original images $I_o \in \mathcal{R}^{\mathcal{H} \times \mathcal{W} \times 3}$ as the input, where $\mathcal{H}$ and $\mathcal{W}$ represent the width and height of the input image, respectively. We use a CNN backbone $f_o(\cdot)$ to extract image feature $f_o(I_o) \in \mathcal{R}^{D_o}$ from $I_o$, where $D_o$ denotes the dimension of the feature channel. Another branch, the HA knowledge branch, incorporates the gaze features of this image. We multiply the gaze information~(HA) with the input image by $I_g = I_o \odot A$, where $A \in \mathcal{R}^{\mathcal{H} \times \mathcal{W} \times 1}$ is the HA saliency map. Through this operation, pixels in the image get different weights from the gaze: the area where humans pay attention to is brighter than the rest. $I_g$ contains visual features which are important for the classification. Another CNN backbone $f_g(\cdot)$ is utilized to extract the gaze feature as $f_g(I_g) \in \mathcal{R}^{D_g}$. Then the gaze feature and original image feature are concatenated together to form the fused feature $f(I_o, I_g) \in \mathcal{R}^{(D_o+D_g)}$. It this way, we integrate HA into a multiclass classification task to study the potential of HA to improve the performance of the image classifier. The whole network is trained with cross-entropy loss.

\section{Experiment}
\label{sec:experiments}
In this section, we first introduce datasets and implementation details. Then we show the results of our proposed GAT and KFN. To show the general validity of our methods, We test on two datasets: CUB-GHA and Eye Gaze Data for Chest X-rays (CXR-Eye)\ \cite{Karargyris2020}. 

\subsection{Datasets and implementation details}
CUB-GHA includes 11788 images in total, with 5994 images for training and 5794 for validation~\cite{WahCUB_200_2011}.
Each image contains eye gaze data from 5 participants. CXR-Eye includes 1083 chest X-ray images with gaze data from a radiologist while performing routine radiology readings \cite{Karargyris2020}. The goal of this dataset is to make a prediction based on the chest X-ray image, whether the subject has one of two clinically prevalent diseases (pneumonia or congestive heart failure (CHF)), or the subject is healthy (normal). The human gaze data is also visualized in the saliency map style. Each image is annotated with one label out of three classes. We choose this dataset because it is a unique human gaze dataset in the medical domain. For such safety-critical applications (e.g. computer-aided diagnosis), we believe the integration of human attention can increase the acceptance and trust of these applications among users.
 
In our experiments on the CUB dataset, the input images are resized to $448 \times 448$ (the images are cropped to this size with the smaller edge first resized to 448) and then randomly flipped horizontally in training. We use the SGD optimizer \cite{ruder2016overview} with an initial learning rate of 0.001. In the experiments on the CXR dataset, the input images are resized to $224 \times 224$ and a random horizontal flip is used in training. We use the Adam optimizer \cite{kingma2014adam} with an initial learning rate of 0.0005. Since the CXR-Eye dataset is relatively small, we run 5-fold cross validation and report the average accuracy of the five validation sets as the final score. All experiments are run for totally 100 epochs training on a single NVIDIA GeForce RTX 3090 and the learning rate decreases after every 50 epochs by a factor of 0.1.

For GAT and KFN, we use ResNet-50 \cite{he2016deep} and EffiecientNet-b5 \cite{tan2019efficientnet} pretrained on ImageNet as backbones on CUB and CXR, respectively. In GAT, we crop the original image using three sets (large, medium and small) of window sizes~(more details can be found in the supplementary material).
Inside each set of window sizes, we run a sliding window algorithm and get $k$ augmentation images for each image in the training set. Concretely, $k$ is set to 2 for large, 3 for medium and 4 for small scale, which results in 9 augmentation images in total. When combining GAT and KFN, we use the GAT trained classifier as backbone in our KFN and fine-tune the KFN for only 20 epochs. 


\subsection{Evaluation on CUB-GHA}
\myparagraph{Ablation study.} To measure the influence of GAT and KFN on the fine-grained classification, we design an ablation study on the CUB dataset where we train a ResNet-50 with cross-entropy loss as the baseline, and several variants by adding GAT and KFN training modules to the baseline. 
From the results shown in Table \ref{tab:cub ablation}, we observe that both GAT and KFN can improve the fine-grained classification accuracy by a large margin. GAT (with HA) improves the baseline model by 2.42\% to 88\%, which indicates that human gaze falls on areas containing discriminative features for classification. When using HA in KFN, the accuracy score is increased from 85.58\% to 86.99\%, which demonstrates that KFN integrates the knowledge of human attention successfully. To show the effectiveness and uniqueness of HA knowledge, we use two machine explanation methods Grad-CAM~\cite{20_grad_cam} and IG~\cite{IG} 
as the saliency maps, replacing HA in GAT and KFN. HA surpasses both methods in the GAT and KFN modules, e.g. KFN~(HA) gains $86.99\%$ while KFN~(IG) gains $85.66\%$. It indicates that human gaze contains unique knowledge that can not be acquired by the model itself. From the result of GAT+KFN, we observe that the combination of both exceeds using any of them alone. 

\begin{minipage}{\linewidth}
  \begin{minipage}[]{0.4\linewidth}
    \vspace*{0.4cm}
    \centering
    \resizebox{\linewidth}{!}{%
        \begin{tabular}{c|c|c}
        \hline
        \multicolumn{2}{c|}{Method}  & Acc. \\
        \hline\hline
\multicolumn{2}{c|}{ResNet-50\ \cite{he2016deep}} &  {85.58} \\ \hline
\multirow{3}{*}{GAT} 
& Grad-CAM\ \cite{20_grad_cam}  & 87.68   \\ 
& IG\ \cite{IG} &  87.73  \\ 
& HA     & 88.00   \\ \hline
\multirow{3}{*}{KFN} 
& Grad-CAM\ \cite{20_grad_cam}  &  85.04    \\ 
\multicolumn{1}{c|}{} & IG\ \cite{IG}  &  85.66   \\ 
\multicolumn{1}{c|}{}  & HA    & 86.99   \\ \hline
GAT+KFN & HA   & \textbf{88.66} \\ \hline

        \end{tabular}
        }
        \vspace{0.2cm}
         \captionof{table}{Ablations study of GAT and KFN on CUB. ``Acc." denotes the accuracy in \%.}
         \label{tab:cub ablation}
\end{minipage}
  \hspace*{1.0cm} 
  \begin{minipage}[]{0.45\linewidth}
    \centering
    \vspace{-0.2cm}
    \resizebox{.7\linewidth}{!}{%
        \begin{tabular}{c|c}
        \hline
        Method  & Acc.  \\
        \hline\hline
        MixUp  \cite{zhang2017mixup}  & 86.23  \\
        CutMix \cite{yun2019cutmix}  & 86.15  \\
        SnapMix \cite{huang2020snapmix}  & 87.75 \\
        Ours (GAT)   & \textbf{88.00}  \\
        \hline\hline
        OSME+MAMC\cite{sun2018multi}  & 86.30  \\
        TASN \cite{zheng2019looking} & 87.90 \\
        API \cite{zhuang2020learning}   & 87.70 \\
        ACNet \cite{ji2020attention}  & 88.10 \\
        Ours (KFN+GAT) & \textbf{88.66} \\\hline
        
        \end{tabular}
        }
 \vspace{0.5cm}
\captionof{table}{Comparison with the state-of-the-art methods on CUB. \textbf{Top:} Comparison of GAT with data augmentation methods. \textbf{Bottom:} Comparison of GAT+KFN with attention-based models.}
    \label{tab:cub sota}
    \vspace*{-0.1cm}
 \end{minipage}
  \end{minipage}



\myparagraph{Comparison with state-of-the-art.} 
We compare our proposed modules with several state-of-the-art methods. Note that for a fair comparison, we compare with the results of using ResNet-50 as the backbone and the input resolution of $448 \times 448$. First, we compare our GAT with other data augmentation methods, i.e., MixUp \cite{zhang2017mixup}, CutMix \cite{yun2019cutmix} and SnapMix \cite{huang2020snapmix} in Table \ref{tab:cub sota} (top). 
The difference between our GAT and other data augmentation methods is that we do not generate synthetic images. MixUp combines two images and their labels linearly, while the rest replace one part of the image with one part from other images. Our GAT simply extends the dataset with the cropped images, which introduces very low computation cost to train the classifier. Among all these works, training a ResNet-50 with GAT outperforms with other state-of-the-art augmentation methods and achieves an accuracy of $88\%$. Moreover, this better trained backbone can be combined easily with other framework to further improve the performance, for instance we combine it with our KFN and thus get better results. 

\begin{table}[b]
\vspace{-0.3cm}
\begin{center}
\resizebox{\textwidth}{!}{
\begin{tabular}{|l|c|c|c|c|c|c|}
\hline
Method &  S3N~\cite{ding2019selective} & S3N + GAT (Ours) & CrossX~\cite{luowei@19iccv}&  CrossX + GAT (Ours) & MMAL~\cite{zhang2021multi} & MMAL + GAT (Ours)\\
\hline
Accuracy & 87.95\% & 88.91\% & 87.70\% &  88.51\%  & 89.25\% & \textbf{89.53}\%\\
\hline
\end{tabular}
}
\end{center}
\caption{Combining our GAT model with the state-of-the-art methods on CUB.}
\label{tab:combi}
\end{table}

We compare our full network with the attention-based methods on CUB in Table \ref{tab:cub sota} (bottom). We choose these methods (OSME+MAMC \cite{sun2018multi}, TASN \cite{zheng2019looking}, API \cite{zhuang2020learning} and ACNet \cite{ji2020attention}) due to their high performance and relevance in simulating human attention by attention modules. They apply attention modules to capture discriminative features from the intermediate output in the network, while we use and integrate the HA directly. 
For instance, \cite{sun2018multi,ji2020attention} applies several layers on the top of the output of the residual block to obtain the region features; API \cite{zhuang2020learning} simulates the comparison behavior of humans as our participants do in the data collection in order to learn discriminative representations. 
Our full network outperforms all state-of-the-art models, achieving $88.66\%$ compared to the attention networks API~($87.70\%$) and ACNet~($88.10\%$). The high performance of our KFN and GAT validates that human gaze can benefit a model's performance in the task.

We combine our module with other state-of-the-art models flexibly and thus improve the performance. In Table \ref{tab:combi}, we show our re-implementations with official code and our improvement by combining our GAT in S3N \cite{ding2019selective}, CrossX \cite{luowei@19iccv} and MMAL \cite{zhang2021multi} models. Please note that no HA information is needed in the inference phase. Our combination of MMAL and GAT improves MMAL from 89.25\% to 89.53\%. We improve CrossX from 87.70\% to 88.51\% and S3N from 87.95\% to 88.91\%, which also surpass the best results given in \cite{ding2019selective,luowei@19iccv}.

\begin{figure*}
    \centering
    \includegraphics[width=\linewidth]{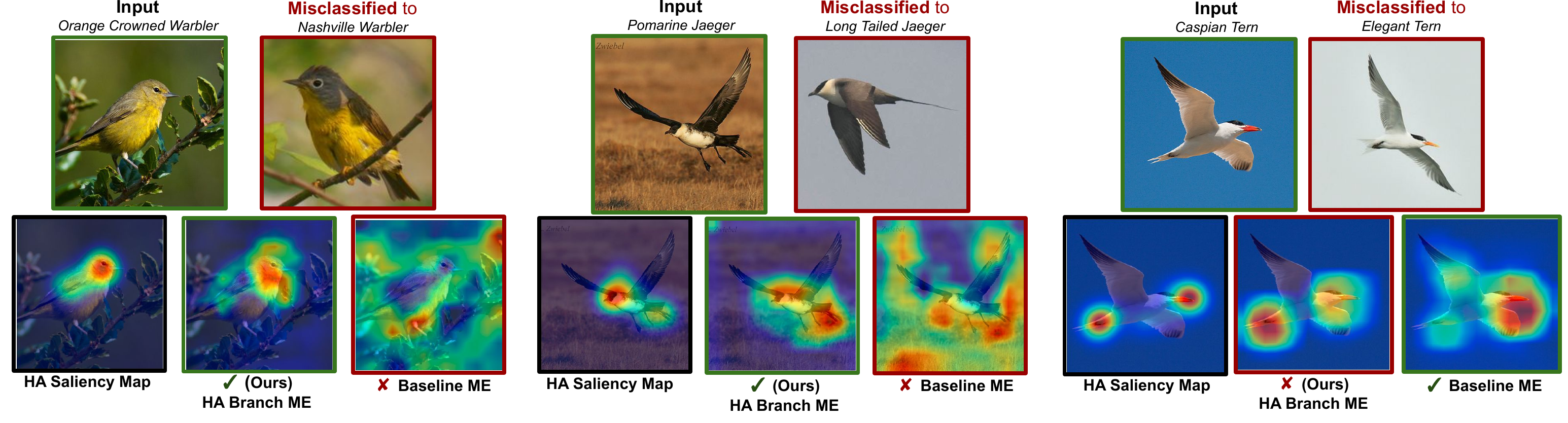}
    \vspace{0.1cm}
    \caption{Illustration of model explanations using HA. Two improved examples and one failure example of our model are shown. For each example, we show the input and misclassification classes; HA saliency map, model explanation of our model, and the baseline model.}

    \label{fig:improvment qualitative}
\end{figure*}

\myparagraph{Qualitative results.} We show two examples from two classes whose accuracy is improved the most compared to the baseline model (vanilla ResNet-50), and one example of a class where our model fails to classify correctly  in Figure \ref{fig:improvment qualitative}. In the first example, the baseline model looks at the belly of an Orange Crowned Warbler and misclassifies it as a Nashville Warbler who also has a yellow fluffy belly. Our model instead focuses on the throat, which is discriminative between the two classes: an Orange Crowned Warbler has a yellow throat, while a Nashville Warbler has a clear mixture of gray and yellow colors on its throat. In the second example, the discriminative feature is the tail. The baseline model mistakes the background as the tail, while our model localizes the tail successfully. Moreover, our model explanation is also more compact and similar to the human saliency map. In the third example, we show a failure of our model: Our model attends to the feet instead of beak which causes the misclassification of a Caspian Tern as an Elegant Tern. Although our model aligns with the human attention, it puts more weight on the feet of birds, since the color of feet is an important feature for distinguishing between a Caspian Tern and a Common Tern (or an Artic Tern).

\subsection{Evaluation on CXR-Eye}
\myparagraph{Comparison with state-of-the-art.}
The state-of-the-art work on CXR-Eye \cite{karargyris2021creation} uses the Efficient-b5 \cite{tan2019efficientnet} as the classifier, however, it deploys random splits to create training, validation and test sets. For a fair comparison, we re-run its network using our 5-fold cross validation setting and report the average of five validation accuracies as the score for this method. The result of this baseline is 70.97\%. When implementing GAT, the result is improved to 71.86\%; when implementing KFN, the accuracy is improved by 3.45\% to 74.42\%. The full model (GAT+KFN) achieves 75.35\% exceeding Efficient-b5 \cite{karargyris2021creation} by 4.38\%. When comparing the performance boost from GAT and KFN, the KFN improves the model on CUB more than GAT. The reason for the difference is how the gaze data is collected. 

\begin{wrapfigure}{r}{0.45\textwidth}
  \begin{center}
   \vspace*{-0.5cm}
    \includegraphics[width=0.45\textwidth]{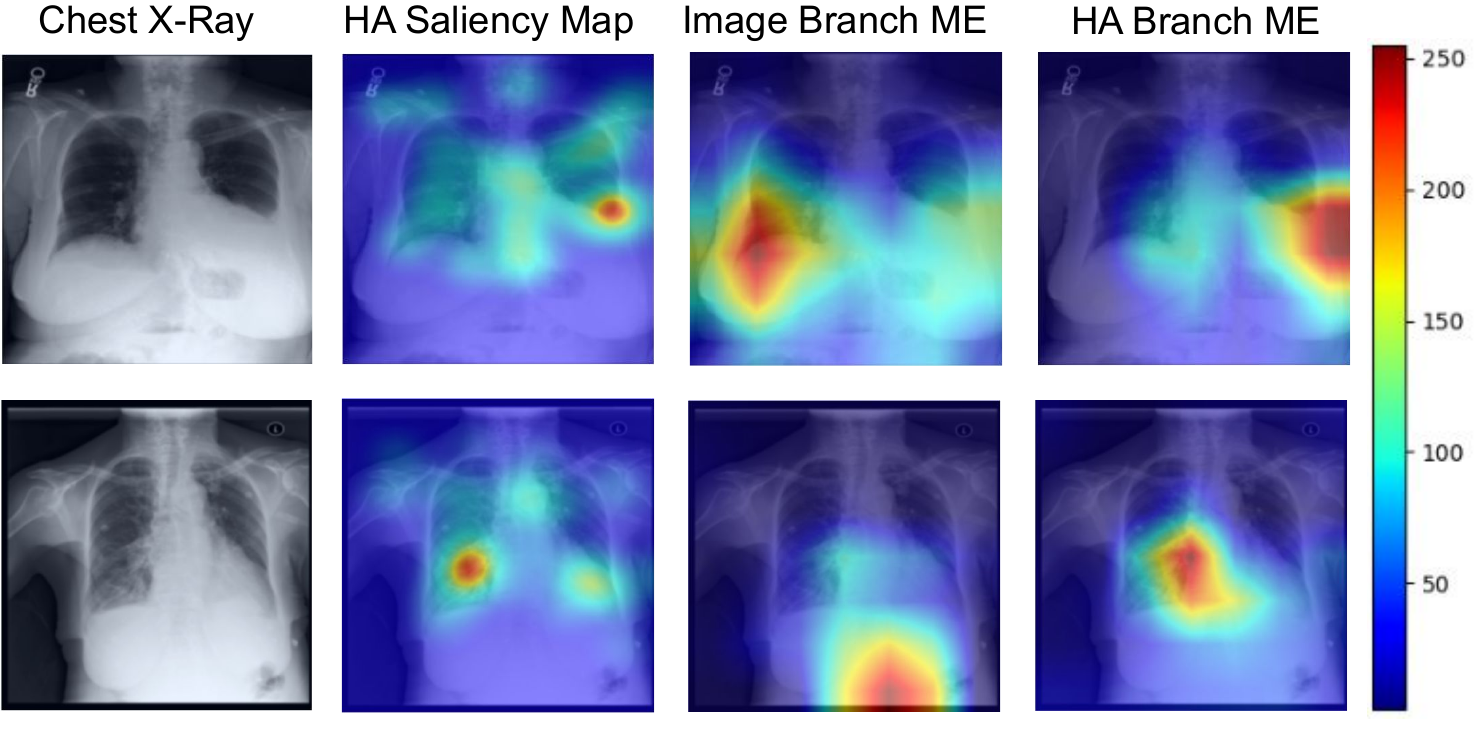}
  \end{center}
  \caption{Illustration of the influence of using HA in model explanation. \textbf{Left to Right:} the original Chest X-ray image; HA saliency map; Model explanation of the Image Branch (w/o HA knowledge) and Model explanation of the HA Branch. }
  \label{fig:cxr}
   \vspace*{-.2cm}
\end{wrapfigure}

In CXR-Eye, the gaze data of the radiologist is collected in an interpretation routine. From the examples shown in Figure \ref{fig:cxr} (sec. column), we see that fixations spread over many locations (light blue area). These locations may play an important role in diagnoses, but GAT localizes the area that the radiologist fixates for relatively longer time. KFN can integrate the knowledge of all potential locations therefore improves the performance by a larger margin.

\myparagraph{Qualitative results.}
To study the influence of integrating HA into the network, we compare the model explanation (Grad-CAM \cite{20_grad_cam}) of each branch in KFN and the qualitative results are shown in Figure \ref{fig:cxr}. From the figure, we see that the HA branch follows more the human attention while the image branch is focusing different areas. 

In the first example (top), human attention focuses more on the left side than the right and the HA branch also does, while the image branch looks more on the right side. The image branch in the second example concentrates on a wrong area, but the HA branch corrects the attentive area to the right. Therefore, KFN improves the performance compared to a model only using images. Most importantly, incorporating gaze knowledge helps to increase the trust and acceptance of the model-based decision in applications such as medical diagnostics, since the model aligns with human behavior.

\section{Conclusion}
\label{sec:conclusion}
In this work, we investigate human attention in classification tasks on the CUB and CXR datasets. In particular, we collect a new gaze dataset, CUB-GHA, and show that human attention focuses on the discriminative regions for a fine-grained classification task. To study the hypothesis that human attention helps a model in the decision-making, we propose the Gaze Augmentation Training and Knowledge Fusion Network which integrate human attention knowledge into the network. Our proposed method improves the accuracy in classification by a large margin on both datasets, showing the general validity of our methods. Thus, our work indicates that human attention provides hints on distinct features in different classification tasks. 

The aim of our work is to demonstrate the potential benefit of human gaze data in classification. As a by-product of this work, we provide the research community with a gaze-enriched dataset CUB-GHA, which can be incorporated with other existing comprehensive annotations (textual explanations, attributes and bounding boxes, etc.). Researchers can therefore validate multiple applications, where human gaze is required in the interaction with a machine.

\section{Acknowledgement}
This work has been partially funded by the ERC (853489 - DEXIM) and by the DFG (2064/1 – Project number 390727645). The authors thank all participants who contributed to the CUB-GHA dataset.

\bibliography{egbib}

\pagebreak
\begin{center}
\textbf{\large Supplemental Materials}
\end{center}
\setcounter{equation}{0}
\setcounter{figure}{0}
\setcounter{table}{0}
\setcounter{section}{0}
\makeatletter
\renewcommand{\theequation}{S\arabic{equation}}
\renewcommand{\thefigure}{S\arabic{figure}}
\renewcommand{\thetable}{S\arabic{table}}
\renewcommand{\bibnumfmt}[1]{[S#1]}
In this document, we provide technical details about our data collection and experiments. First, we explain how we set the standard deviation of the Gaussian distribution in the Human Attention (HA) saliency map generation and show more analyses on gaze data including the relationship between human fixation points and the discriminative attributes of birds. In addition, quantitative and qualitative comparisons between the model explanations (MEs) and HA are demonstrated. In the second section, we introduce implementation details (e.g. sliding window sizes) in the Gaze Augmentation Training (GAT). 

\section{CUB-GHA}
\subsection{HA Saliency Map Generation}
\begin{figure}[h]
    \centering
    \includegraphics{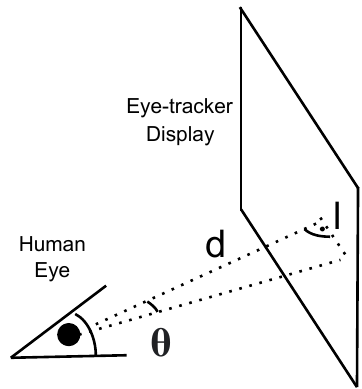} 
    \vspace{0.1cm}
    \caption{Illustration of a human observing an image on the eye-tracker display.}
    \label{fig:gaussian}
\end{figure}

Figure \ \ref{fig:gaussian} illustrates a human observing an image on the eye-tracker display. As mentioned in the paper, we post process every fixation location as a Gaussian distribution ${N}(\mu,\,\sigma^{2})$ on the HA saliency map, where $\sigma$ is 75 pixels (in the display's resolution). We calculate the standard deviation $\sigma$ as follows. 
In our experiment setup, the distance $d$ between the human eye and the eye-tracker display is 60 $cm$, and the visual angle $\theta$ is set to 2$^{\circ}$ following \cite{vickers2007perception}. In this case, $l=\tan{2^{\circ}} \cdot d = 21$ $mm$. According to the settings of display, in the horizontal direction the length of the display is 530 $mm$ and the resolution is 1920 pixels. Therefore, we can get that $l=$ 21 $mm$ covers approximately 75 pixels on the display. We set 75 pixels as the standard deviation with the image rescaled to the display resolution (1920 $\times$ 1080). The saliency map is rescaled to its original size afterwards.

\subsection{Gaze Data Analysis}

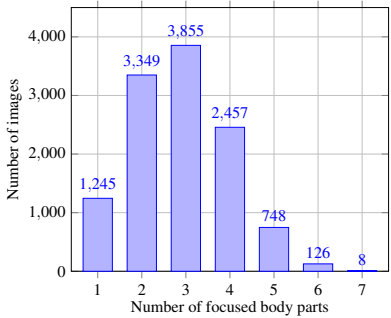
\begin{figure}[t]
    \centering
    \resizebox{.4\linewidth}{!}{%
    \begin{tikzpicture}
\begin{axis}[
        xlabel={Number of focused body parts},
        ylabel={Number of images},
        set layers,
        ybar=1.2pt,
        bar width=18pt,
        symbolic x coords={1,2,3,4,5,6,7},
        grid=both,
        ymin=0,
        legend cell align=left,
        ymax=4500,
        x label style={},
        y label style={},
        ticklabel style={},
        nodes near coords,
        ]
    \addplot coordinates {
        (1,1245) 
        (2,3349) 
        (3,3855) 
        (4,2457)
        (5,748)
        (6,126)
        (7,8)
    };
    \end{axis}
    \end{tikzpicture}
    }
    \vspace{0.2cm}
    \caption{Histogram of the number of focused bird body parts in CUB-GHA. \textbf{Y-axis} refers to the amount of images with the certain number of parts (\textbf{X-axis}).}
    \label{fig:bar chart}
\end{figure}

In this section, we validate that the attributes discovered by our collected human gaze data are discriminative for the fine-grained classification. CUB includes ground-truth attributes for each image and they are 312-dimension binary vectors.
We use them to conduct the ground-truth discriminative attributes of each bird class in the dataset. There are 100 comparison pairs in the data collection experiments, and we compare each image from the first class with every image in the second class. For instance, if there are $M$ images in the first class and $N$ in the second one, there are in total $M\cdot N$ combinations between the two classes. For each combination, we conduct a comparison attribute vector where $1$ is set if that attribute entry is the same for both images, or $0$ if not the same, i.e. the comparison attribute vector is also a 312-dimension binary vector. We sum $M\cdot N$ comparison attribute vectors together to have one 312-dimension vector representing the ground-truth discriminative attribute for these two classes. For instance if the attribute \texttt{has-wing-color::brown} in the comparison vector is 354, it means that the attribute \texttt{has-wing-color::brown} differs in the 354 image pairs. In the end, we group the attributes into seven body parts (head, beak, breast, belly, back, wing, leg). For example, we sum up all the attribute values in the comparison vector that are related to the wing, and the sum represents the difference of the wing between the given two classes. The body part with the highest sum is the most discriminative body part between the two classes.

When our participants look at the image, they always focus on the discriminative body parts of the bird. The body part which human gaze falls in should contain the largest number of different attributes between the compared two classes. With the help of body part center coordinates in each image, we can assign every fixation (collected for this image from five participants) to its nearest body part according to the distance between the center coordinate and the fixation coordinate. In Figure \ref{fig:bar chart}, we show the histogram of the number of focused bird body parts on the whole CUB-GHA dataset. We see that there are three body parts focused by humans in 3855 images. Most of the images (92.52\%) include less than five parts focused in the dataset. In very few images, our participants view all seven parts of the bird. In each image, we sum up the duration of fixations belonging to one body part and use it to represent the amount of human attention on that part. A longer duration sum indicates more attention participants have paid. We rank the seven body parts for each image according to the duration sums and calculate the rate that the top-$k$ focused body parts hit the most discriminative one (which is conducted from the ground-truth attributes). The hit rate is shown in Table \ref{tab:hit rate}.
From the results, we see that our participants discover the most discriminative body part in 84.4\% of the images correctly. Within four parts that participants consider to be important for the classification, the ground-truth distinct body part is found in 98.3\% of the images. This result shows that human gaze data in CUB-GHA hints on discriminative body parts/attributes in the classification.

\begin{table}[h]
\centering
\begin{tabular}{|l|l|l|l|l|}
\hline
Top-k         & 1     & 2     & 3     & 4     \\ \hline\hline
Hit rate (\%) & 84.40 & 93.60 & 97.18 & 98.31 \\ \hline
\end{tabular}
\vspace{0.5cm}
\caption{Hit rate of the most discriminative body part. Top-$k$ refers to the $k$ longest focused body parts by humans in CUB-GHA.}
\label{tab:hit rate}
\end{table}

\subsection{Comparison between ME and HA}
In this section, we provide more details and results of comparing MEs and HA. We use the KAR (keep and retrain) procedure \cite{hooker2018benchmark} and the concrete procedure works as follows: given an input image $I \in \mathcal{R}^{\mathcal{H} \times \mathcal{W} \times 3}$ and the importance estimation map $A \in \mathcal{R}^{\mathcal{H} \times \mathcal{W} \times 1}$, where $H$ and $W$ represent the width and height of input image, respectively; $A$ can be the HA or ME saliency map. We construct a mask $M \in \mathcal{R}^{\mathcal{H} \times \mathcal{W} \times 1}$ to filter the pixels in $I$. First, we sort $A$ in a descending order to $A^R$ according to the attention values. Then we binarize $A$ by taking the top $p$ percent of pixels in $A^R$ as one and others as zero:
\begin{equation*}
    M(x,y)=\begin{cases}1.0, & \text{if }  (x,y) \in P \\0.0, & \text{otherwise} \end{cases} \,,
\end{equation*}
where $P$ are the indices of top ranked $p$ percent pixels.
We apply the mask $M$ to filter the corresponding image $I$ in the training and testing set: $I' = M \odot I$, so that only the top $p$ percent of the most important features are observed by the network. After such a modification of the dataset, we train a new model and compare the test accuracy. This procedure aims at evaluating whether the important feature estimated by $A$ (i.e. model or human attention) is critical to the classification or not. A good estimation $A$ encodes important features in a small amount of pixels. In other words, a higher accuracy with such small amount of pixels indicates that the given features are more important. We generate the new dataset using an insertion percentage $p$ = [5,10,15,20,15,30,50,70,90] and train the vanille ResNet-50 \cite{he2016deep} using the same hyper-parameters as in the baseline training. We run this procedure three times independently from random initialization for each estimation map and report the average accuracy on the test set. 

\begin{figure*}[t]
\centering
         \begin{center}
            \includegraphics[width=.75\linewidth]{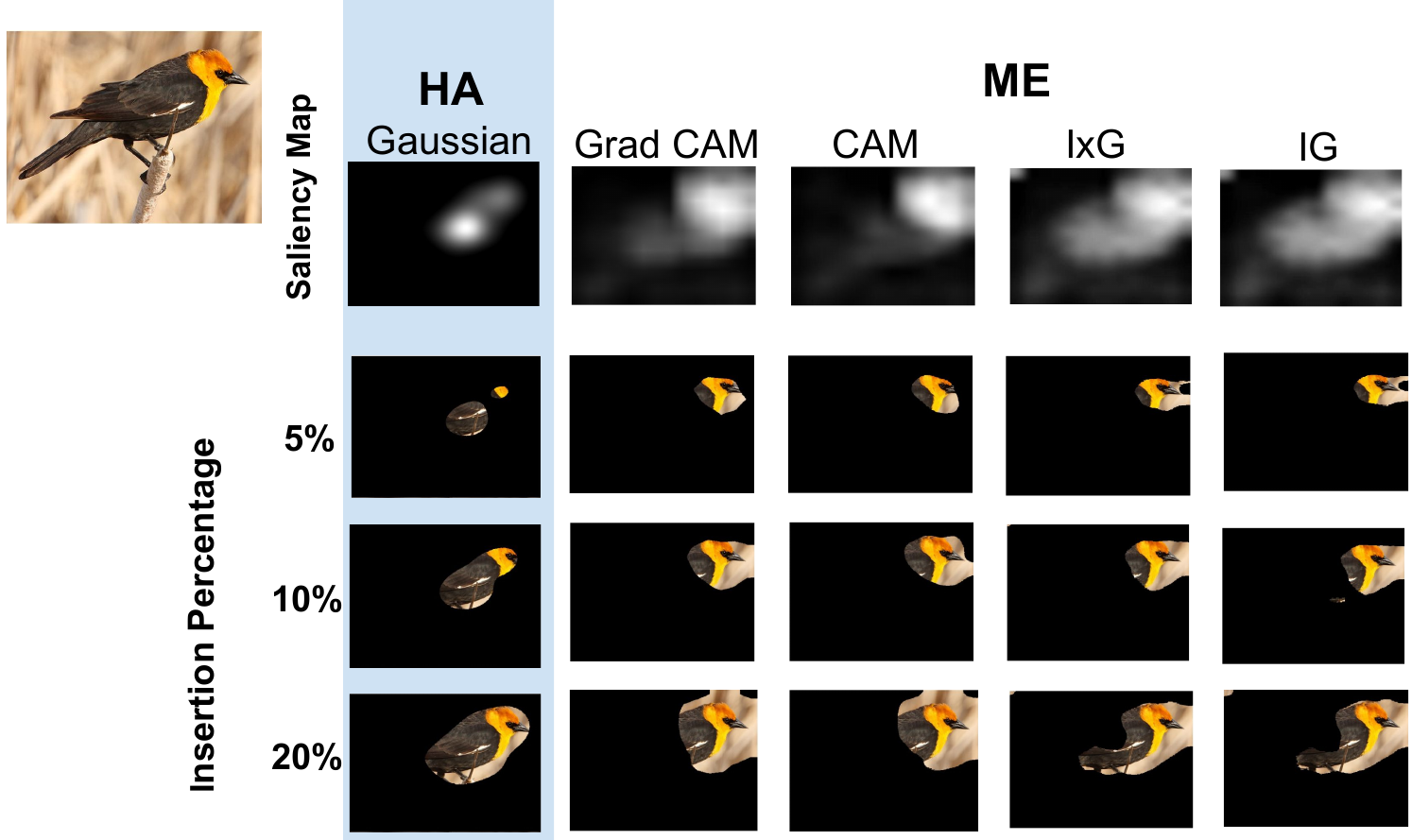}
         \end{center}
   \caption{Modified images in the Keep and Retrain procedure. The pixels are inserted according to the importance in the estimation maps. \textbf{Top to bottom}: importance estimation maps (saliency maps), modified images using top 5\%, 10\% and 20\% important pixels in saliency maps. \textbf{Left to right}: HA, ME-Gradient-based CAM (Grad-CAM)~\cite{20_grad_cam}, Class Activations Maps (CAM)~\cite{19_cam}, InputXGradient (IxG)~\cite{shrikumar2016not} and Integrated Gradients (IG)~\cite{IG}.}
   \vspace{-0.5cm}
\label{fig:illustration reveal}
\end{figure*}

Figure \ref{fig:illustration reveal} illustrates the qualitative results of the modified images using HA and MEs. These differences can be observed when using 5\% and 10\% as insertion percentages. If we compare HA and MEs, they focus on a similar area after 20\% pixels are inserted: the wing and head parts. When comparing among MEs, CAM and IxG are similar to Grad-CAM and IG, respectively. In this example, Grad-CAM/CAM pays more attention to the head, while IG/IxG focuses more on the body. From the qualitative comparison results, we see that the HA and MEs estimate different parts of the bird as being the most important ones for the classification task, especially regarding the first 10\% important regions. 

We also conduct a quantitative similarity comparison between HA and MEs. We evaluate on different metrics: Kullback-Leibler divergence (KL-D), correlation coefficient (CC) and similarity (SIM), which are often used in comparisons of how similar two images are \cite{bylinskii2018different}; rank-correlation (Rank-Co) as introduced in \cite{das2017human}; shuffled AUC metric (sAUC) evaluating every pixel in saliency maps as a classification task; information gain (IG) measuring the performance over a baseline \cite{bylinskii2018different, qiuxia2020understanding}. CAM is very similar to Grad-CAM, e.g. Grad-CAM achieving 0.565 on CC and 1.242 on KL-D, while CAM achieving 0.563 and 1.248, respectively. Additionally, we observe IG and IxG achieving similar performances on these metrics, i.e. 0.699 for IG v.s. 0.694 for IxG on CC, and 1.318 for IG v.s. 1.310 for IxG on KL-D. These similarities can be seen from the qualitative results as well. From all different metrics, we see that the Grad-CAM tends to be the most similar to HA, as Grad-CAM achieves the highest scores in all six metrics. This is consistent with the results from the KAR that Grad-CAM achieves the best performance among all MEs.

\setlength{\tabcolsep}{3pt}
\renewcommand{\arraystretch}{1.1}
\begin{table}[h]	
	\centering
	\resizebox{0.6\linewidth}{!}{
    \begin{tabular}{l|cccccc}
            & KL-D $\downarrow$ & CC $\uparrow$  & SIM $\uparrow$ & Rank-Co $\uparrow$  & sAUC $\uparrow$ & IG $\uparrow$ \\
            \hline

    CAM  & 1.248 & 0.563 & 0.399  & \textbf{0.761}  & 0.460 &   0.938         \\
    Grad-CAM  & \textbf{1.242}  & \textbf{0.565} & \textbf{0.415} &  \textbf{0.761} & \textbf{0.508} & \textbf{1.376} \\
    IG  & 1.318 & 0.546 & 0.361 & 0.699  & 0.436 &  0.921       \\
    IxG  & 1.310  & 0.543 & 0.375 & 0.694 & 0.461 & 1.001\\
    \end{tabular}
    }
    \vspace{0.4cm}
    \caption{Similarity comparison between MEs and HA saliency map. 
    ($\downarrow$: the lower the better; $\uparrow$: the higher the better.)
    }
    \label{tab:quantitative}
\end{table}

\section{GAT Experiments}
\begin{table}[h]
\centering
\resizebox{\linewidth}{!}{%
\begin{tabular}{c|c|c|c}
\hline
     & Small    & Medium    & Large    \\ \hline
CUB-GHA  & (123,134) (134,123) (123,123) (134,134)  & (174,190) (190,174) (174,174) (190,190) & (246,264) (269,246) \\ \hline
CXR-Eye & (87,95) (95,87) (95,95) (87,87) &  (123,135) (135,123) (123,123) (135,135) & (180,190) (190,180)  \\\hline
\end{tabular}
}
\vspace{0.5cm}
\caption{Sliding window size used in GAT.}
\label{tab:window size}
\end{table}

\begin{figure}[h]
    \centering
    \includegraphics[scale=0.4]{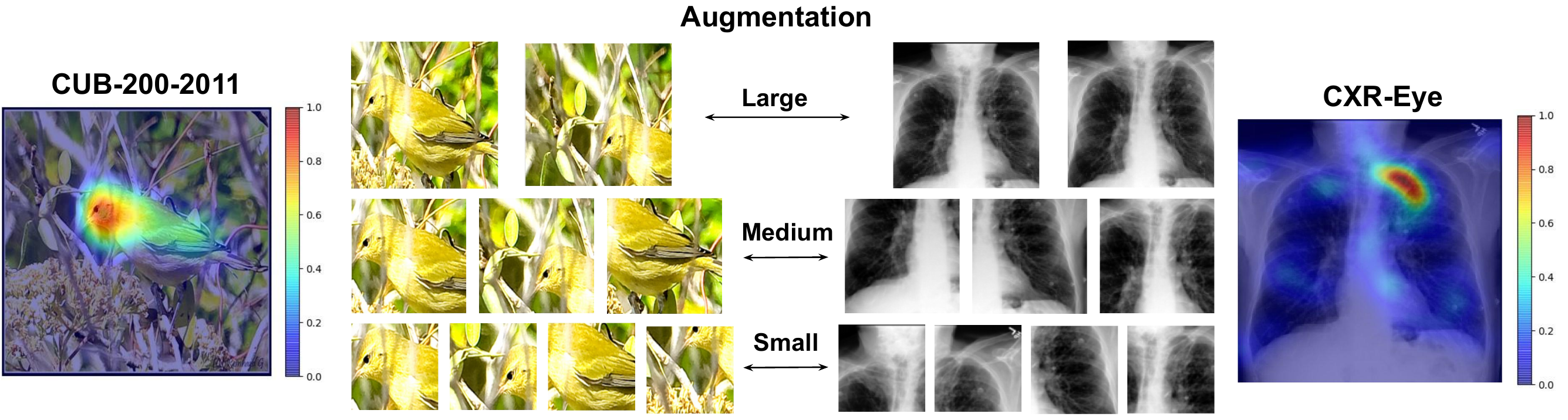}
     \vspace*{0.2cm}
    \caption{Illustration of cropped images used in the Gaze Augmentation Training. \textbf{Left and Right:} HA saliency maps used for augmentation on CUB-GHA and CXR-Eye. \textbf{Middle:} cropped images in three scales (large, medium and small).}
    \label{fig:GAT illustration}
\end{figure}

Concrete sliding windows sizes $(w,h)$ used for each dataset in GAT experiments are listed in Table \ref{tab:window size}. For the CUB-GHA dataset, we choose the sliding window sizes based on the averaged size of bird bounding boxes: the width is 246 and the height is 269 if images are resized to $448 \times 448$. Therefore, we use 246 and 269 as sizes for the large scale. The medium window size is conducted using the factor of $\frac{\sqrt{2}}{2}$ to have the half of the bounding box area, i.e. we use 174 and 190 as window size options. The factor used in the small scale is $0.5$. For the CXR-Eye dataset, we choose $0.8$ and $0.85$ as factors with respect to the resized image size $224 \times 224$ for the large window size, i.e. two options are 180 and 190. Similarly, factors for the medium window size are $0.55$ and $0.6$. The small window sizes are scaled based on the medium window sizes by the factor of $\frac{\sqrt{2}}{2}$. The motivation of using different sliding window sizes is to get different parts which are discriminative for the classification. To avoid very similar cropped areas, we choose $0.25$ as the iou threshold in the non-maximum suppression. Table \ref{tab:window k} lists the ablation study of using different numbers of cropped areas ($k$) in the augmentation training on two datasets. (2,2,2) denotes that two cropped areas are picked up from each window scale to form the augmentation training set. We choose (2,3,4) as the final setting since it gives relatively better results on both datasets. Figure \ref{fig:GAT illustration} illustrates the augmentation images using the setting (2,3,4) in three sets of window scales on both datasets.

\begin{table}[h]
    \centering
        \begin{tabular}{c|c|c}
        \hline
         (L,M,S)  & CUB (\%) & CXR (\%) \\
        \hline\hline
        (2,2,2) & 87.50  & 71.03 \\
        (2,3,2) & 88.06 & 71.58 \\
        (2,3,3) & 88.00 & 71.86\\
        (2,3,4) & 88.00  & 72.21 \\
        \hline
        \end{tabular}
        \vspace*{0.5cm}
        \caption{Results of using different window size settings on CUB-GHA and CXR-Eye. The number of windows used in large, medium and small size is shown on the left. The accuracy is in \%.}
        \label{tab:window k}
\end{table}

\end{document}